УДК 004.8

# Алгоритмы автоматического выделения интентов и классификации высказываний для целеориентированных диалоговых систем

# Algorithms for automatic intents extraction and utterances classification for goal-oriented dialogue systems

Л.В. Легашев[1], silentgir@gmail.com
ORCID: https://orcid.org/0000-0001-6351-404X

А.Е. Шухман[1], shukhman@gmail.com
ORCID: https://orcid.org/0000-0002-4303-2550

А.Ю. Жигалов[1], leroy137.artur@gmail.com
ORCID: https://orcid.org/0000-0003-3208-1629

[1] Оренбургский государственный университет

L.V. Legashev[1], silentgir@gmail.com
A.E. Shukhman[1], shukhman@gmail.com
A.Yu. Zhigalov[1], leroy137.artur@gmail.com

[1] Orenburg State University, 13, Pobedi Ave., Orenburg, 460018, Russia

**Аннотация**
Современные методы машинного обучения в домене обработки естественного языка могут использоваться для автоматической генерации сценариев для целеориентированных диалоговых систем. В рамках текущей статьи представлена общая схема исследования автоматической генерации сценариев для целеориентированных диалоговых систем. Описан метод предварительной обработки наборов диалоговых данных в формате JSON. Выполнено сравнение двух методов выделения намерений пользователей на основе BERTopic и латентного размещения Дирихле. Выполнено сравнение двух реализованных алгоритмов классификации высказываний пользователей целеориентированной диалоговой системы на основе логистической регрессии и моделей трансформеров BERT. Подход на основе трансформеров BERT с использованием модели bert-base-uncased показал лучшие результаты по трём метрикам Precision (0,80), F1-score (0,78) и Matthews correlation coefficient (0,74) в сравнении с другими методами.

**Ключевые слова:** целеориентированные диалоговые системы; интенты, высказывания, машинное обучение; BERT; регрессия.

**Введение**

Важнейшим практически значимым направлением искусственного интеллекта в последнее время стала обработка естественных языков NLP (Natural Language Processing). Одним из наиболее популярных направлений исследований является разработка диалоговых систем, способных вести диалог с пользователями, автоматически решая множество рутинных задач, связанных с уточнением информации, ответами на наиболее часто встречающиеся вопросы, решением конкретных проблем пользователя, и т.д. В настоящее время сценарии для диалоговых систем разрабатываются вручную. В то же время на основе имеющихся диалогов возможна автоматическая генерация сценариев для целеориентированных диалоговых систем с использованием современных методов машинного обучения.

Основная научная проблема исследования заключается в снижении детерминированности и ограниченности диалоговых систем за счет использования подхода к автоматической генерации возможных сценариев реакций на высказывания пользователей на основе размеченных диалоговых данных в определенной предметной области. Общая схема исследования представлена на рисунке 1.

На первом этапе необходимо выполнить предварительную обработку диалоговых данных. Для заданного набора диалогов строится множество намерений (intents set) (категорий намерений) пользователей, в состав которого добавляются общие (повседневные) намерения, включающие в себя интенты приветствия, прощания, согласия и отказа с репликами оператора, удовлетворенностью работы и т.д. Также строится множество сценарных блоков (scenario blocks set), в состав которого будут входить стандартные и наиболее часто используемые блоки ответов на высказывания пользователей. Среди таких блоков можно выделить следующие:

— предоставление ответа на запрос пользователя;
— ответ на общие интенты пользователя (приветствие, прощание, согласие, отказ);
— запрос информации из базы данных по теме пользователя (topic:database request);
— добавление внешнего источника данных, в случае отсутствия информации (no_info: topic);
— и другие.

Будет выполнена разметка данных диалогов в соответствии с множеством намерений и множеством сценарных блоков.

Далее необходимо реализовать методы для выделения тематик намерений пользователей по их высказываниям. Для решения этой задачи могут быть использованы методы как классического тематического моделирования, так и нейросетевого тематического моделирования, основанные на получении векторных представлений текстов глубокими нейронными сетями на архитектуре трансформеров. Аналогично разрабатываются алгоритмы для выделения сценарных блоков ответов операторов.

После этого требуется разработать алгоритмы классификации высказываний по интентам, которые должны обучаться на размеченных данных диалогов.

Для решения задачи автоматической генерации графа сценария будет построено дерево принятия решений, в котором вершинами будут намерения пользователей на каждом этапе диалога, а дугами - возможные вариации ответов оператора, в зависимости от категории намерения и предметной области. Планируется использовать эвристический алгоритм поиска Монте Карло, который позволит выделить несколько взвешенных переходов к ответам операторов на основе обученных диалоговых данных выбранной прикладной области.

В рамках задачи также планируется построить множество пар в формате ключ:значения, где каждой категории намерения пользователя будут соответствовать несколько ранжированных ответов оператора от наиболее вероятного, к менее вероятному. Если пользователь не будет доволен исходом диалога, для предпоследнего обработанного интента будет выбран другой сценарный блок, в результате чего дальнейший диалог будет перестроен. В рамках алгоритма можно будет зафиксировать общее число итераций повторения диалога в зависимости от размера множества значений ранжированных пар ответов оператора.

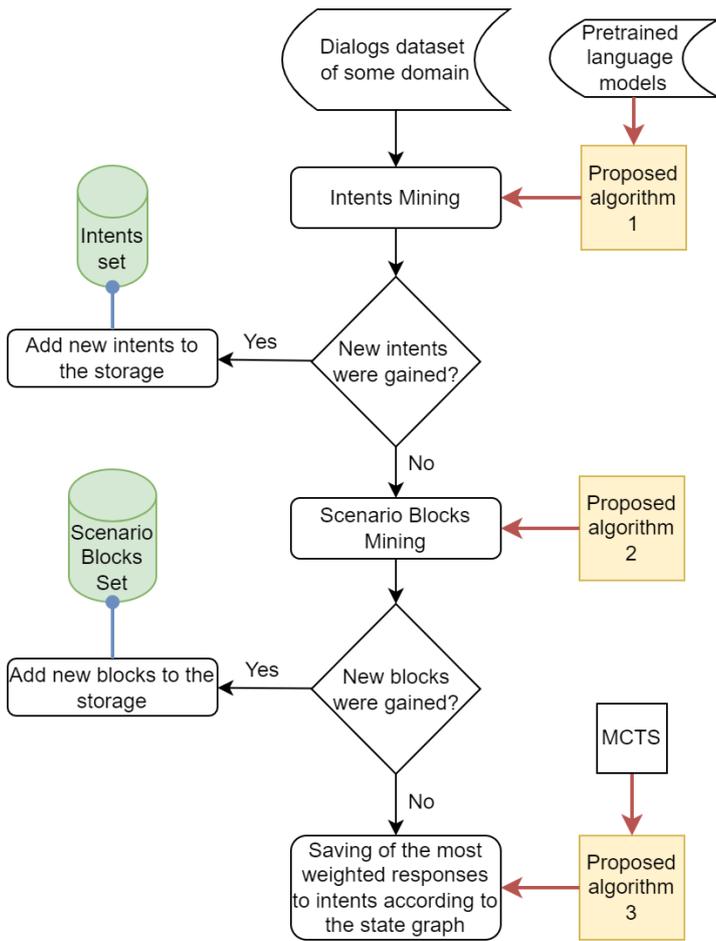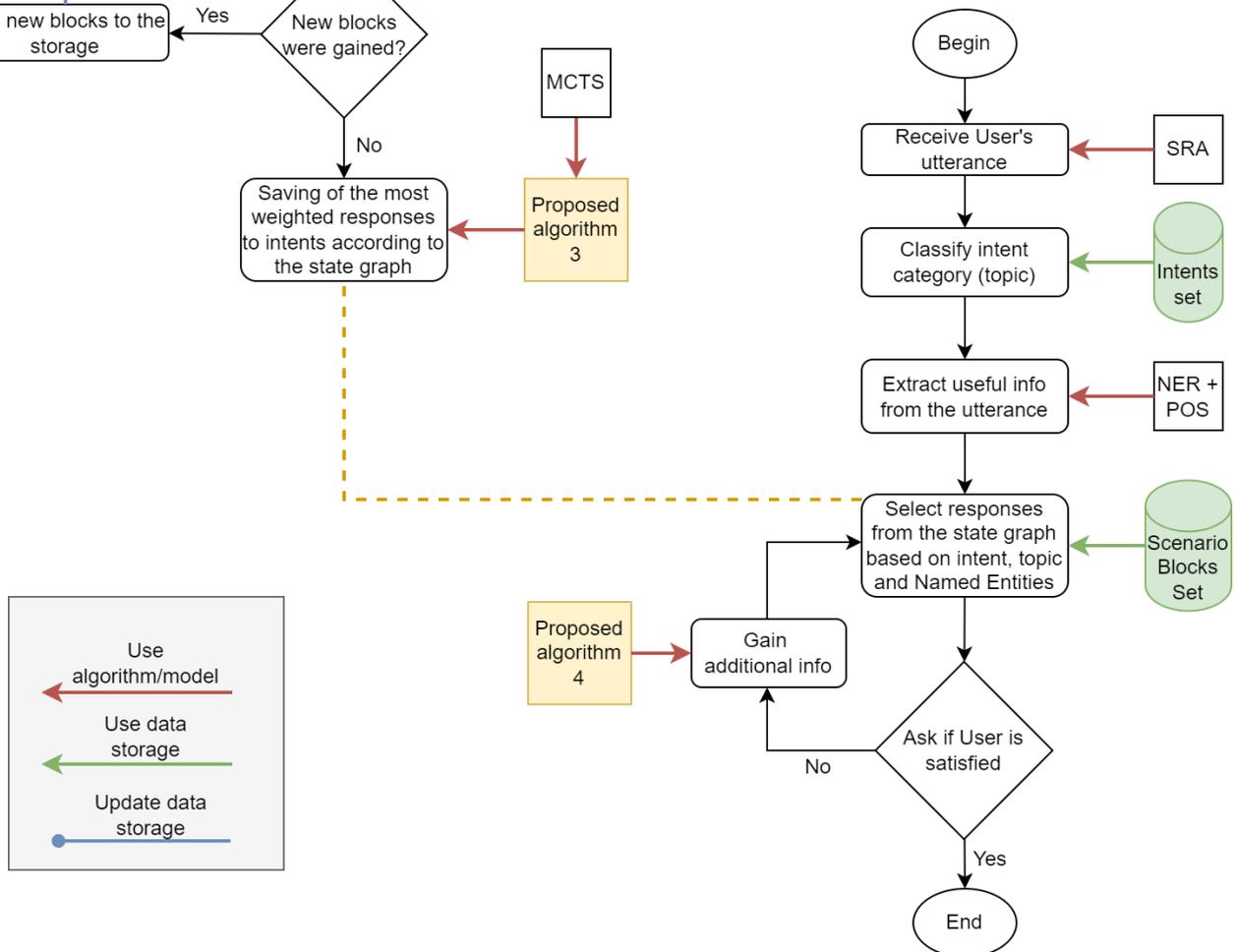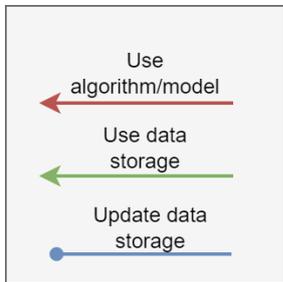

Рисунок 1 – Общая схема исследования автоматической генерации сценариев для целеориентированных диалоговых систем с поддержкой непрерывного обучения

Figure 1 - General scheme for the study of automatic scenario generation for goal-oriented dialogue systems with continuous learning support

В некоторых случаях от пользователя потребуется запросить дополнительную информацию в соответствии с его высказыванием и выбранной категорией интента. Будет выполнена оценка информативности высказывания (information content of the utterance) на основе заранее выработанных правил. Например, если пользователь желает забронировать отель, но не указывает в своем высказывании точную дату и количество человек, оператор дополнительно запрашивает информацию вида date:checkin, date:checkout, adults:number и т.д.

Для актуализации алгоритмов генерации и преодоления дрейфа модели будет реализовано непрерывное машинного обучение, использующее содержательные метаданные с автоматической переоценкой весов моделей. На основе сохранения и анализа диалогов в процессе эксплуатации системы планируется проводить дообучение моделей, включая выделение новых намерений и сценарных блоков, коррекции графа сценария и модели контекста диалога.

В исследовании будут использованы методы классификации текстов на основе предварительно обученных глубоких нейронных сетей, методы нейросетевого тематического моделирования, глубокие генеративные модели на архитектуре трансформеров, алгоритмы выделения именованных сущностей. В рамках текущей статьи будут представлены методы предварительной обработки наборов диалоговых данных, выделения тем на основе неструктурированных документов и классификации высказываний целеориентированной диалоговой системы.

**1 Обзор исследований**

На текущий момент времени проблема автоматического формирования сценариев для целеориентированных диалоговых систем является новой для исследований в области обработки естественных языков. При этом множество исследований связаны с генерацией ответов на вопросы пользователя в форме осмысленного диалога. В работе [1] представлен условный вариативный фреймворк для генерации диалогов, на основе полуавтоматической глубокой генеративной модели. В работе [2] описывается бенчмарк генерации естественного языка для симуляции настроек обучения на малом количестве данных в проблемно-ориентированных диалоговых системах. С целью улучшения генерации ответов в диалоговых системах авторы статьи [3] предлагают фреймворк, который использует один и тот же кодировщик для извлечения общих и не зависящих от проблемы признаков с помощью разных декодеров для изучения признаков, специфичных для проблемы. В статье [4] предлагается новый фреймворк, который использует обучение с подкреплением для улучшения качества генерации диалогов в проблемно-ориентированных диалоговых системах. Авторы исследования [5] генерируют ответы на основе извлечения знаний из неструктурированных документов для проблемно-ориентированных диалоговых систем. В работе [6] представлен фреймворк многоцелевой аугментации данных, который позволяет генерировать множество разнообразных подходящих ответов диалога, которые приводят к достижению цели диалога.

Ключевой задачей при разработке диалоговых систем является задача распознавания намерений (intents, интенты) пользователей и их классификация. Авторы статьи [7] используют датасет обслуживания клиентов, обучая модели без учителя представления текста и разрабатывания модель отображения намерений пользователей, которая ранжирует намерения на основе косинусного сходства между предложениями и намерениями. В работе [8] используется архитектура двунаправленного энкодера и механизмы на основе внимания для для распознавания намерений. Авторы исследования [9] генерируют тренировочный датасет интентов с помощью предложенного основанного на плотности алгоритма кластеризации. В статье [10] проведено исследование того, данные какого объема необходимы для задачи классификации интентов. Авторы делают вывод о том, что 25 тренировочных экземпляров на каждую категорию намерений достаточно чтобы достичь доли правильных ответов в 94%.

Эвристический алгоритм поиска Монте Карло (Monte Carlo Tree Search, MCTS) используется для принятия решений в классах задач, для которых на каждом ходу существует множество возможных исходов. Указанный алгоритм можно адаптировать для решения задачи построения диалоговой системы, в которой для каждого намерения пользователя можно подобрать один или несколько взвешенных переходов к ответу системы. Небольшая часть исследований посвящена этому вопросу. В статье [11] алгоритм MCTS используется в связке с моделями глубокого обучения для решения задачи построения диалога при бронировании билетов в кино. В работе [12] представлен Байес-адаптивный алгоритм планирования для целеориентированных диалоговых систем с использованием рекуррентных нейронных сетей и алгоритма MCTS. В публикации [13] алгоритм MCTS используется для оценки качества сгенерированной последовательности диалогов в состязательных сетях, генерирующих последовательности диалоговых систем. В статье [14] автор выполняет поиск по пространству состояний ответов в диалогах малой и средней длины с использованием алгоритма MCTS.

Нейросетевое тематическое моделирование (neural topic modeling) используется для кластеризации произвольных текстовых корпусов по "темам" и может применяться для задач выделения намерений пользователей и сценарных диалоговых блоков. В работе [15] представлена техника нейросетевого тематического моделирования BERTTopic, которая генерирует векторные представления документов с помощью предварительно обученных языковых моделей и формирует представления тем с помощью процедуры TF-IDF (Term Frequency, Inverse Document Frequency) на основе классов. В статье [16] оценивают производительность четырех методов тематического моделирования: скрытого распределения Дирихле (LDA), неотрицательной матричной факторизации (NMF), Top2Vec и BERTopic при анализе данных социальной сети Twitter.

## 2 Метод предварительной обработки наборов диалоговых данных

Для решения задачи предварительной обработки наборов диалоговых данных мы рассмотрели англоязычный датасет MultiWOZ 2.2 (Multi-Domain Wizard-of-Oz) [17], содержащий письменные разговоры между людьми на тематику поиска и бронирования отелей, ресторанов, поездов, достопримечательностей и т.д. Представленные данные содержат поочередный диалог между пользователем (USER) и системой (SYSTEM), каждый шаг диалога (turn) представляет собой одно высказывание (utterance) пользователя или системы (например, '*I'm looking for a local place to dine in the centre that serves chinese food.*'). Суммарное количество диалогов в наборе данных MultiWOZ 2.2 составляет 8 438 с суммарным количеством шагов в 42 190.

Данные хранятся в формате json, на первом этапе метода мы объединяем все json файлы в один с помощью операции конкатенации. На следующем этапе мы используем три операции для извлечения интересующих нас данных, а именно само высказывание, намерение (intent) и автор высказывания (speaker). Операции и подробное описание каждой из них представлены в таблице 1.

**Таблица 1. Операции извлечения данных из наборов диалоговых систем формата JSON.**
**Table 1. Operations for data extraction from dialog systems datasets of JSON format.**

| Операция | Описание |
| --- | --- |
| df['turns'][i][j]['speaker'] | Автор высказывания на *j*-ом шагу *i*-ого диалога |
| df['turns'][i][j]['utterance'] | Текст высказывания на *j*-ом шагу *i*-ого диалога |
| df['turns'][i][j]['frames'][0]['state']['active_intent'] | Интент высказывания на *j*-ом шагу *i*-ого диалога |

Полученный после выполнения операций датасет содержит два признака – высказывание и намерение ('sentence','intent'), при этом категории интентов и их количество представлены в таблице 2. Удалим из итогового датасета записи с интентами поиска госпиталя (find_hospital) и поиска автобуса (find_bus) по причине их малого количества. Итоговый датасет с девятью классами назовем USER_INTENTS.

**Таблица 2. Категории намерений в наборе данных, отсортированные по убыванию.**
**Table 2. Intent categories in dataset sorted in descending order.**

| Категория интента | Описание интента | Количество записей |
|---|---|---|
| NONE | Реплики общего характера | 18560 |
| find_restaurant | Найти ресторан | 9393 |
| find_hotel | Найти гостиницу | 7954 |
| find_attraction | Найти достопримечательности, экскурсии, выставки | 4694 |
| find_train | Найти поезд | 4580 |
| book_restaurant | Забронировать места в ресторане | 3974 |
| book_hotel | Забронировать номера в гостинице | 3452 |
| find_taxi | Заказать такси | 2839 |
| book_train | Забронировать билеты на поезд | 1109 |
| find_hospital | Найти медицинское учреждение | 214 |
| find_bus | Найти общественный транспорт | 7 |

В категорию NONE полученного датасета попали все промежуточные реплики пользователей во время диалогов. Необходимо построить множество намерений (intents set) (категорий намерений) пользователей, в состав которого будут добавлены общие (повседневные) намерения, включающие в себя интенты приветствия, прощания, согласия и отказа с репликами оператора, удовлетворенностью работы и т.д. В следующем разделе будет показано, как можно использовать метод BERTopic для выделения новых категорий намерений в тексте.

## 3 Метод выделения намерений пользователей по их высказываниям для целеориентированной диалоговой системы

Тематическое моделирование используется в области обработки естественного языка для нахождения скрытых тематик в документах с целью определения принадлежности произвольного документа к той или иной теме или кластеру. В данном случае речь идет о машинном обучении без учителя, т.к. модель обучается на неразмеченных данных. Основная идея тематического моделирования заключается в подсчете слов и группировке словесных паттернов (структур) для определения тем в неструктурированных текстовых документах. Наиболее популярными моделями тематического моделирования в машинном обучении являются вероятностный латентно-семантический анализ PLSA (Probabilistic Latent Semantic Analysis) и латентное размещение Дирихле LDA (Latent Dirichlet Allocation), в котором предполагается, что распределение тематик которой имеет распределение Дирихле.

Метод выделения намерений пользователей будет основан на технике нейросетевого тематического моделирования BERTTopic, которая генерирует векторные представления документов с помощью предварительно обученных языковых моделей и формирует представления тем с помощью процедуры TF-IDF на основе классов. Отфильтруем данные из полученного датасета USER_INTENTS, оставив только записи с категорией намерений NONE. Запустим технику тематического моделирования BERTopic с моделью эмбеддингов paraphrase-MiniLM-L12-v2 и зададим минимальный размер темы min_topic_size=50. Пятнадцать наиболее популярных тем и 5 ключевых слов представлены на рисунке 2.

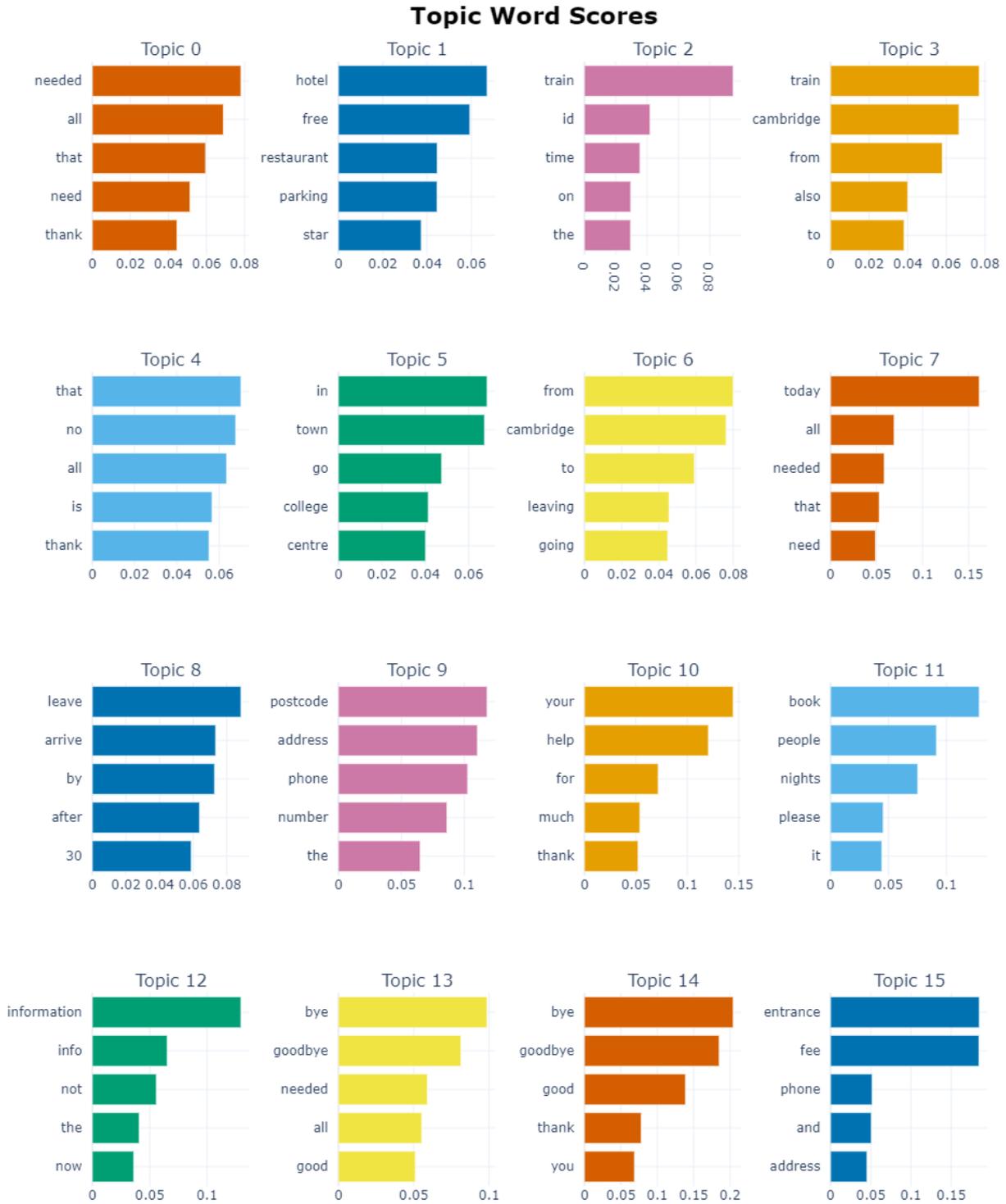

Рисунок 2. Выделение тем в высказываниях пользователей с помощью BERTTopic.

Figure 2. Topics mining in user utterances using BERTTopic.

В дополнение к девяти категориям интентов из таблицы 2 можно добавить следующие новые категории интентов на основе анализа выделенных тем: пользователь удовлетворен и прощается (satisfied_goodbye), пользователь выражает благодарность (gratitude), пользователь выбирает время прибытия поезда (set_arrival_time), пользователь запрашивает номер телефона и индекс (phone_number_postcode), пользователь запрашивает стоимость билетов (entrance_fee), пользователь ищет полицейский участок (find_police_station).

Выполним сравнение выделенных пятнадцати тем методом BERTopic с работой метода LDA. Результаты представлены в таблице 3. На основе анализа тем можно выделить обнаруженные ранее интенты satisfied_goodbye, gratitude, phone_number_postcode и set_arrival_time. При сравнении двух методов следует отметить более корректную группировку ключевых слов в темах при использовании метода BERTopic.

Таблица 3. Первые 15 тем, полученные методом латентного размещения Дирихле.
Table 3. The first 15 topics obtained by Latent Dirichlet Allocation method.

| Тема | Ключевые слова |
|---|---|
| Topic: 0 | that, thank, all, you, be, no, will, thanks, for, much |
| Topic: 1 | to, train, on, cambridge, need, after, by, 15, and, leaving |
| Topic: 2 | the, time, you, need, travel, that, what, will, be, no |
| Topic: 3 | for, need, the, you, is, hospital, please, that, police, yes |
| Topic: 4 | all, that, thanks, today, needed, thank, you, bye, goodbye, no |
| Topic: 5 | for, you, need, that, there, the, to, do, taxi, great |
| Topic: 6 | to, am, looking, also, find, place, for, stay, you, help |
| Topic: 7 | to, from, like, cambridge, on, would, also, train, should, and |
| Topic: 8 | you, that, is, need, thank, all, no, for, very, thanks |
| Topic: 9 | you, for, thank, that, book, great, will, yes, it, no |
| Topic: 10 | the, and, have, free, please, with, parking, just, for, need |
| Topic: 11 | your, help, thank, you, for, much, so, all, that, day |
| Topic: 12 | you, for, thank, is, that, need, it, no, the, help |
| Topic: 13 | in, the, of, town, same, as, to, area, and, go |
| Topic: 14 | can, number, please, me, phone, and, you, the, address, get |

**4 Алгоритмы классификации высказываний целеориентированной диалоговой системы**

В рамках текущего раздела решается задача многоклассовой (всего 9 категорий интентов) классификации намерений пользователей для произвольного англоязычного текста. Для решения этой задачи будут рассмотрены два подхода. Первый подход заключается в использовании логистической регрессии, которая основана на линейном разделении объектов и преобразует текст в его векторное представление. Второй подход заключается в использовании англоязычных моделей трансформеров BERT, который основан на дообучении предварительно обученной нейронной сети с дополненными слоями классификатора на размеченном наборе данных.

4.1 Логистическая регрессия

На первом этапе выполняется числовое кодирование целевой переменной – названия девяти категорий высказываний. Задается словарь стоп-слов из англоязычного корпуса биб-лиотеки `nltk` и задается минимальная и максимальная длина *n*-грам от 1 до 3. На следующем этапе объект CountVectorizer использует модель мешка слов (bag-of-words), формируя словарь *n*-грамм длины *m*, и каждый текст представляется вектором длины *m*, в котором каждый элемент соответствует количеству вхождений соответствующей *n*-граммы в текст. Объект TfidfVectorizer вместо количества вхождений для *n*-граммы сохраняет коэффициент TF-IDF.

4.2 Модели трансформеры BERT

На первом этапе также выполняется числовое кодирование целевой переменной. Задается максимальный размер словаря num_words = 15000 и максимальная длина высказывания пользователя max_len = 200 в токенах, происходит выравнивание предложений исходного датасета до одинаковой длины (padding='post'). На следующем этапе выполняется токенизация обучающей выборки с помощью моделей bert-base-uncased и xlm-roberta-base. Векторные представления формируются с помощью входного слоя нейронной сети на основе списка словарных номеров текстовых токенов. Функция softmax библиотеки torch используется для получения предсказанной вероятности принадлежности выборки к одной из девяти категорий высказываний.

4.3 Сравнение подходов и оценка моделей

Результаты сравнения двух исследуемых подходов классификации высказываний пользователей по метрикам Precision, F1-score и Matthews correlation coefficient (MCC) представлены в таблице 3.

**Таблица 3. Сравнение метрик для исследуемых подходов.**
**Table 3. Comparison of metrics for the approaches under study.**

| Подход | Precision | F1-score | MCC |
|---|---|---|---|
| Logistic Regression and CountVectorizer | 0,7440 | 0,7393 | 0,6806 |
| Logistic Regression and TfidfVectorizer | 0,7535 | 0,7508 | 0,6951 |
| **BERT with bert-base-uncased model** | **0,7962** | **0,7798** | **0,7392** |
| BERT with xlm-roberta-base model | 0,7716 | 0,7602 | 0,7159 |

Подход на основе трансформеров BERT с использованием модели bert-base-uncased показал лучшие результаты по всем трём метрикам в сравнении с другими подходами. В таблице 4 представлено несколько примеров классификации интентов для произвольно написанных англоязычных текстов для модели bert-base-uncased model.

**Таблица 4. Примеры классификации высказываний пользователей.**
**Table 4. Examples of user utterances classification.**

| Пример текста | Интент |
|---|---|
| 'I am looking for a train that will leave after 14:00 and should depart from Cambridge.' | find_train |
| 'I would like to visit some portraits exhibition, Van Gogh maybe?' | find_attraction |
| 'My name is Alex. Can we have a conversation?' | NONE |
| 'Hi, can you recommend me some place with Mexican food and drinks?' | find_restaurant |
| 'I would like to book a train to Sacramento, today, 5 p.m. one ticket please.' | book_train |
| 'Good evening. I need a taxi to the airport. This is urgent!' | find_taxi |
| 'This hospital has excellent facilities. I was satisfied with the treatment.' | NONE |
| 'Please call me a taxi to drive to the central area, I would like to have a dinner in Italian restaurant at 5 p.m.' | find_restaurant |

Из последнего примера видно, что в одном предложении пользователя содержатся несколько возможных интентов. Если мы выведем три наиболее вероятных интента, то получим следующие результаты: find_restaurant: 38 %, book_restaurant: 29 % и find_taxi: 31 %. Выделение множества интентов и их ранжирование может быть полезным для диалоговой системы с целью определения дальнейшего сценария построения беседы с пользователем.

**Заключение**

Таким образом, в рамках данного исследования представлена общая схема исследования автоматической генерации сценариев для целеориентированных диалоговых систем. Описан метод предварительной обработки наборов диалоговых данных на примере англоязычного датасета MultiWOZ 2.2 в формате JSON. Выполнено сравнение двух методов выделения намерений пользователей – BERTopic и латентное размещение Дирихле – что позволило выделить несколько новых интентов в исходном датасете. Выполнено сравнение двух алгоритмов классификации высказываний пользователей целеориентированной диалоговой системы на основе логистической регрессии и моделей трансформеров BERT. Подход на основе трансформеров BERT с использованием модели bert-base-uncased показал лучшие результаты по трём метрикам Precision (0,80), F1-score (0,78) и Matthews correlation coefficient (0,74) в сравнении с другими методами. В рамках дальнейших исследований планируется реализация метода выделения сценарных блоков и разработка модели генерации графа сценариев для целеориентированной диалоговой системы в произвольной прикладной области с сохранением контекста диалога.